\documentclass[journal]{IEEEtran}
\usepackage{amsmath}
\usepackage{lineno}
\usepackage{soul}
\usepackage{cite}
\usepackage{url}

\usepackage{booktabs}
\usepackage{multirow}
\usepackage[table,xcdraw]{xcolor}
\definecolor{NewColor1}{HTML}{A0A2ED}
\usepackage{placeins}
\usepackage{paralist}
\usepackage[inline]{enumitem}
\usepackage[vlined,linesnumbered,ruled,resetcount]{algorithm2e}
\newcommand{\nextnr}{\stepcounter{AlgoLine}\ShowLn}
\usepackage{bm}
\usepackage{blindtext}
\usepackage{tikz}
\usetikzlibrary{positioning,spy}
\usepackage{breqn}

\usepackage{float}
\usepackage[caption = false]{subfig}

\makeatletter
\newcommand{\clearsubcaptcounter}{\setcounter{sub\@captype}{0}}
\makeatother

\usepackage{etoolbox, siunitx}

\usepackage{amssymb}
\usepackage{pifont}
\ifCLASSINFOpdf
\else
\fi
\begin{document}
%
\title{Segmentation Guided Deep HDR Deghosting}
%
%
%

\author{K~Ram~Prabhakar,
        Susmit~Agrawal,
        and~R~Venkatesh~Babu,~\IEEEmembership{Senior~Member,~IEEE}
\thanks{K.~R.~Prabhakar, S.~Agrawal and R.~V.~Babu are with the Department of Computational and Data Sciences, Indian Institute of Science, Bengaluru, KA, 560012, INDIA. e-mail: ramprabhakar@iisc.ac.in.}
}

%
%

\markboth{Journal of \LaTeX\ Class Files,~Vol.~14, No.~8, August~2015}%
{Shell \MakeLowercase{\textit{et al.}}: Bare Demo of IEEEtran.cls for IEEE Journals}
%



\maketitle

\begin{abstract}
We present a motion segmentation guided convolutional neural network (CNN) approach for high dynamic range (HDR) image deghosting. First, we segment the moving regions in the input sequence using a CNN. Then, we merge static and moving regions separately with different fusion networks and combine fused features to generate the final ghost-free HDR image. Our motion segmentation guided HDR fusion approach offers significant advantages over existing HDR deghosting methods. First, by segmenting the input sequence into static and moving regions, our proposed approach learns effective fusion rules for various challenging saturation and motion types. Second, we introduce a novel memory network that accumulates the necessary features required to generate plausible details in the saturated regions. The proposed method outperforms nine existing state-of-the-art methods on two publicly available datasets and generates visually pleasing ghost-free HDR results. We also present a large-scale motion segmentation dataset of 3683 varying exposure images to benefit the research community. The code and dataset are available at \url{http://val.serc.iisc.ernet.in/HDR/shdr/}.
\end{abstract}


%
\IEEEpeerreviewmaketitle

\section{Introduction}
    \label{sec:intro}
\IEEEPARstart{H}{uman} eyes can recognize a wide range of illumination present in nature, a range significantly higher than any standard digital camera's capability. Our eyes can discern details in both highlight and shadow areas of the scene. However, cameras have difficulty in capturing both ends of the brightness spectrum. The captured image will likely have highlight or shadow regions, where some parts are correctly exposed while a few details will be missing in the other. High Dynamic Range (HDR) imaging is a photography technique used to generate images with a wide range of illumination than possible with digital cameras. HDR imaging faithfully reconstructs details in both highlight and shadow regions as perceived by our human eyes.

Using a specialized hardware devices or sensors (\nolinebreak\hspace{1sp}\cite{tocci2011versatile,mcguire2007optical,zhao2015unbounded}) is one way to generate HDR image. However, such a custom device is costly for widespread consumer use. Instead, a widely adopted way is to shoot a stack of bracketed varying exposure low dynamic range (LDR) images to capture both ends of the brightness spectra. The exposure stack is then merged to generate a single HDR image with well-exposed highlights and shadows. Many methods (\nolinebreak\hspace{1sp}\cite{debevec2008recovering,granados2010optimal,prabhakar2017deepfuse}) have been proposed in the past to generate HDR from static images. Nevertheless, the images captured in a practical application contain both camera and object motion. Applying such static fusion methods on dynamic scenes result in ghosting artifacts.

\begin{figure*}[ht]
    \centering
    \includegraphics[width=18cm]{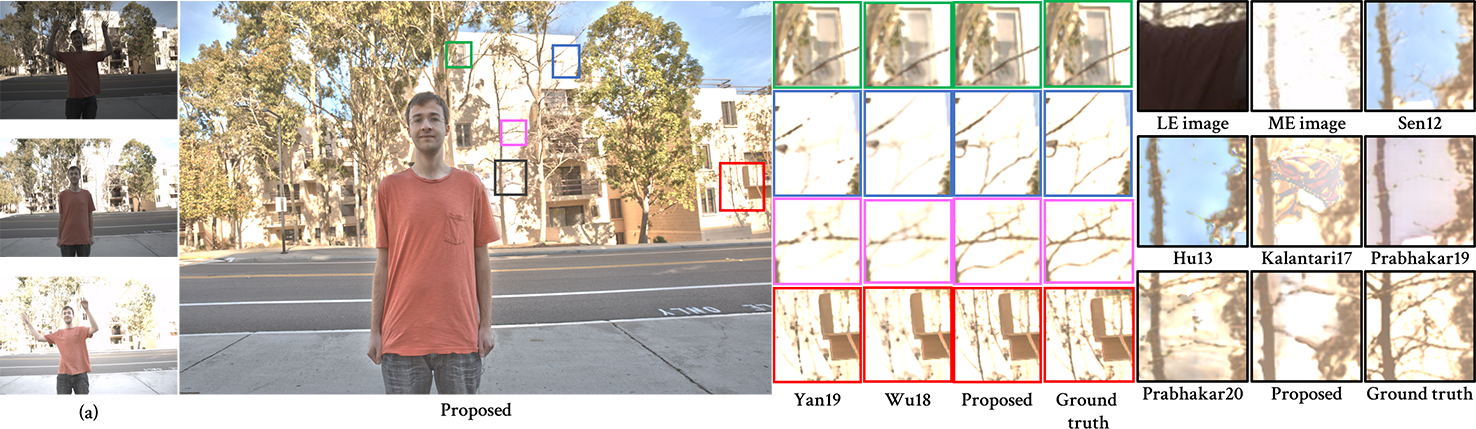}%
    \caption{The input varying exposure images (a) contain gradual exposure change and moving objects. The objective of our proposed method is to combine these input images and generate a visually pleasing HDR image without any ghosting. As shown in zoomed regions on the right, the existing methods produce results with artifacts or unable to recover saturated regions for challenging sequences. In our proposed method, we simplify the problem by dividing the inputs into static and dynamic parts. Then, we fuse them independently to generate the final result. As highlighted in the zoomed regions, our method generates ghosting and artifact-free results even in regions affected by motion and saturation. Image courtesy of Kalantari and Ramamoorthi \cite{kalantari2017deep}. Image best viewed zoomed-in electronic display.}
    \label{fig:result1}
\end{figure*}

HDR deghosting methods in the literature can be classified into four major categories. \begin{inparaenum}[1)]
    \item \textit{Pixel rejection methods}: These methods align the input stack using global alignment techniques. The static pixels across all images are merged using standard static HDR fusion techniques. Whereas, the dynamic pixels from the affected images are rejected from the fusion process (\nolinebreak\hspace{1sp}\cite{khan2006ghost,jacobs2008automatic,pece2010bitmap,zhang2012reference,grosch2006fast,gallo2009artifact,Shanmuga2011}). As these methods reject pixels in the dynamic region, they suffer from having only LDR content in the moving regions.
    \item \textit{Alignment methods}: These methods align images using non-rigid approaches and merge them using simple fusion techniques (\nolinebreak\hspace{1sp}\cite{bogoni2000extending,kang2003high, jinno2008motion,zimmer2011freehand,hu2012exposure,gallo2015locally}). However, as these methods use simple fusion techniques, they cannot handle alignment errors.
    \item \textit{Synthesis methods}: These methods synthesize missing details due to saturation in the reference image from other input images using patch-based optimization techniques (\nolinebreak\hspace{1sp}\cite{hu2013hdr,sen2012robust,ma2017robust}). They generate static sequences from the given dynamic sequence and merge them using standard static fusion methods. These methods introduce artifacts in challenging sequences with heavily saturated regions (see Fig. \ref{fig:result1}), and they are computationally expensive.
    \item \textit{Deep learning methods}: With the rise of deep learning, many approaches have been proposed recently to generate HDR images with Convolutional Neural Networks (CNN) (\nolinebreak\hspace{1sp}\cite{kalantari2017deep,wu2018deep,yan2019attention, prabhakar2019fast,prabhakar2020}). These methods generate significantly better HDR images than traditional non-deep methods. However, these methods still produce artifacts for challenging scenes with heavily saturated regions. As highlighted in examples Fig. \ref{fig:result1}, \ref{fig:result2} and \ref{fig:result4}, the current state-of-the-art methods have artifacts in their results and there is room for improvement in HDR deghosting.
\end{inparaenum}

We propose a novel CNN-based HDR deghosting method to address the aforementioned issues. In our approach, we simplify the problem by segmenting the image into dynamic or moving and static regions. Then, we merge static and dynamic features separately using two different CNN fusion sub-networks. The fused dynamic and static features are combined using a decoder network. Unlike Yan \textit{et al.} \cite{yan2019attention}, we train a CNN segmentation network with ground truth segmentation images for accurate static and dynamic segmentation. Also, different from existing state-of-the-art deep learning methods like \cite{kalantari2017deep,yan2019multi,yan2019attention,wu2018deep,prabhakar2020}, our method can be extended to fuse an arbitrary number of images and achieve better performance compared to Prabhakar \textit{et al.} \cite{prabhakar2019fast} in the same category. Furthermore, we present a novel global memory network to improve the quality of images predicted by model.

The existing deep learning-based methods use a single CNN model to fuse both static and dynamic regions. Conversely, we take inspiration from the literature and follow the divide-and-conquer approach. We divide the given sequence into static and dynamic regions and fuse them separately. However, unlike rejection based methods, we use CNN to learn the efficient fusion rule, thus avoiding the shortcomings of rejection based methods. Our approach combines the divide-and-conquer advantage of rejection methods with better generalization capability of CNNs. In Yan \textit{et al.} \cite{yan2019attention}, the authors use attention mechanisms to select the useful features; however, they do not enforce any loss on the predicted attention maps. Hence, the network is free to choose features that minimize the final HDR loss; may not necessarily identify static and dynamic features (see Fig. \ref{fig:result1}). In our approach, we train a segmentation network dedicated to identifying the dynamic regions and simplify the fusion process. In summary, the main contributions of our work are as follows, 
\begin{itemize}[noitemsep,topsep=0pt,itemsep=0pt]
 \item We propose a motion segmentation assisted CNN for artifact-free HDR deghosting. Our proposed method identifies the dynamic regions and learns different fusion rules for static and dynamic, thus effectively dealing with challenging motion and saturation conditions. 
  \item We propose a new memory network module that enhances the network's ability to generate rich details in heavily saturated regions.
  \item We present a new large-scale motion segmentation dataset with 3683 varying exposure images and their corresponding human annotated motion segmentation maps to benefit the research community.
  \item We perform extensive experiments to demonstrate better performance of the proposed approach. Additionally, through rigorous ablation experiments, we justify the importance of each module in our proposed approach.
\end{itemize}
The rest of the paper is organized as follows. We present the literature methods and their limitations in Section \ref{sec:related}. We discuss our proposed approach elaborately in Section \ref{sec:prop}. Then, we discuss the proposed dataset and present the experimental results in Section \ref{sec:eval}. In Section \ref{sec:discussion}, we discuss the extension of the proposed method to an arbitrary number of images and computational complexity. Finally, we conclude the paper in Section \ref{sec:conc}.
\section{Related Works}
    \label{sec:related}
    HDR deghosting has a vibrant literature history spanning almost two decades. A wide range of algorithms was proposed in the literature to address HDR deghosting. They can be broadly clubbed into four categories.  

The first category of algorithms assumes that input images are mostly static and contains few regions with motion. Hence, they aim to identify dynamic pixels (pixels affected by motion) and process them separately. The static pixels are merged using the conventional HDR merging process. Many different approaches use different techniques to identify dynamic regions. Grosch \cite{grosch2006fast} proposes a method to identify dynamic pixels by thresholding the difference between images. Wu \textit{et al.} \cite{wu2010robust} identify dynamic pixels that violate brightness consistency among the inputs. Gallo \textit{et al.} \cite{gallo2009artifact} threshold the difference between patches in logarithmic domain instead of pixels to identify dynamic pixels. Heo \textit{et al.} \cite{heo2010ghost} estimate the joint probability density function between input images and threshold it to determine the dynamic pixels. Later, they refine it further using an energy minimization formulation. Min \textit{et al.} \cite{min2009histogram} perform multi-level thresholding of intensity histograms between input images to recognize dynamic pixels. Raman \textit{et al.} \cite{raman2011reconstruction} develop on top of \cite{gallo2009artifact} by comparing super-pixels.

Pece \textit{et al.} \cite{pece2010bitmap} generates a median threshold bitmap by thresholding the pixels with median value and use it to find motion regions. Eden \textit{et al.} \cite{eden2006seamless} produce HDR result in two steps. The first step uses a graph cut technique to generate output with a similar structure as that of the reference image but does not have a complete dynamic range in all regions. Then the dynamic range of the output is increased in the second step by borrowing properly exposed details from other input images. Khan \textit{et al.} \cite{khan2006ghost} estimate weight value for each pixel in all input images using two criteria: the probability that it is well exposed and it belongs to the background. Then, images are combined using estimated weight maps. Zhang \textit{et al.} \cite{zhang2012reference} detect dynamic pixels by comparing image gradients. Granados \textit{et al.} \cite{granados2010optimal} propose an optimization method to identify a coherent subset of images that be combined to generate output without ghosting artifacts. 

After identifying dynamic regions, these methods ignore images with dynamic content and use information from the rest of the static images. Whereas, for the static regions, they make use of contents from all images. There are two sub-classes of algorithms within this category. In the first subclass, the dynamic regions are replaced by content from a selected reference image (\nolinebreak\hspace{1sp}\cite{grosch2006fast,min2009histogram,heo2010ghost}). In the second subclass, the dynamic regions are filled by information from static images; thus, these methods will predict static regions in all pixels (\nolinebreak\hspace{1sp}\cite{reinhard2005dynamic,eden2006seamless,khan2006ghost}). These methods are fast and work better for mostly static scenes. Their main drawback, however, is producing LDR content in large moving regions.

The second class of algorithms aligns the input images with a selected reference image to generate a static sequence. Then the conventional HDR merge algorithm is used to fuse them. Ward's \cite{ward2003fast} method aligns images through homography estimated by comparing median threshold bitmaps of the individual images. Tomaszewska \textit{et al.} \cite{tomaszewska2007image} use SIFT features followed by RANSAC to compute homography for aligning images. Bogoni's \cite{bogoni2000extending} method handles camera motion by registering images using a global affine transform and object motion with the optical flow. Kang \textit{et al.} \cite{kang2003high} generate HDR video from alternating exposure sequences. They estimate bidirectional optical flow between the current frame and consecutive frames to align and merge images. Zimmer \textit{et al.} \cite{zimmer2011freehand} propose a optimization-based optical flow method to align images with two energy terms. The data term ensures that images are aligned properly, while the smoothness term ensures smooth flow among neighboring pixels. Gallo \textit{et al.} \cite{gallo2015locally} introduce a fast algorithm suited for real-time mobile processing. Their method speeds up the operation by computing optical flow only at selected sparse locations and interpolating it for the other pixels. Compared to the first class of algorithms, these methods can generate HDR content in moving regions. In general, registration-based methods \cite{ward2003fast,tomaszewska2007image} may fail in case of complex deformable motions. Additionally, optical flow can produce erroneous flow in heavily saturated regions. 

The third class of algorithms synthesizes static sequences from dynamic sequences using patch-match (\nolinebreak\hspace{1sp}\cite{barnes2009patchmatch}) or similar techniques. They begin with selecting one of the input images as a reference. They then synthesize a new sequence with the same structure as the reference but with same exposure as the corresponding image in the input sequence. In the synthesized result, the structure will resemble a reference where the reference is well exposed. Whereas, in poorly exposed regions of reference, the structure is borrowed from other images. Sen \textit{et al.} \cite{sen2012robust} use multi-scale bidirectional similarity metric for locating similar patches. While Hu \textit{et al.} \cite{hu2013hdr} make use of a patch-match algorithm (\nolinebreak\hspace{1sp}\cite{barnes2009patchmatch}) for the same. This class of algorithms can handle complex deformable motion and hallucinate details in the reference saturated regions. While \cite{sen2012robust,hu2013hdr} methods produce high-quality results, they are computationally demanding.

The last class algorithm uses deep learning-based techniques to perform HDR deghosting. Kalantari and Ramamoorthi \cite{kalantari2017deep} begin with aligning images using optical flow. Later, they merge the aligned images using a CNN instead of a conventional HDR merge process. The CNN is trained to merge them by ignoring the artifacts introduced by optical flow. Wu \textit{et al.} \cite{wu2018deep} method proposes a simple CNN-based method to generate the final HDR image directly from misaligned input images. Yan \textit{et al.} \cite{yan2019multi} method uses a multi-scale CNN model to reconstruct the HDR result in a coarse-to-fine strategy. In another work, Yan \textit{et al.} \cite{yan2019attention} propose a method to remove ghosting artifacts using an attention mechanism. Yan \textit{et al.} \cite{yan2020deep} propose a CNN model with non-local module. A non-local module identifies matching neighbor features to fill in the saturated regions in their approach. Prabhakar \textit{et al.} \cite{prabhakar2019fast} introduced a scalable CNN method that can fuse an arbitrary number of images. They achieve scalability by aggregating the mean and max of all input feature maps. Recently, Prabhakar \textit{et al.} \cite{prabhakar2020} proposed a method to process very high-resolution images using Bilateral Guided Upsampler (BGU). While the recent deep learning-based methods have produced better ghost-free results than the traditional methods, their results are still inaccurate in challenging conditions (see Fig. \ref{fig:result1}, \ref{fig:result2}, \ref{fig:result3}, \ref{fig:result4}). 
\section{Proposed method}
    \label{sec:prop}
With $N$ varying exposure LDR images, $\{\mathbf{X}_k\}$, $\forall k=(1,\dots, N)$ as input, the goal of our approach is to merge them into a single HDR image ($\mathbf{Y}$) without any ghosting artifacts.   

\textbf{Motivation}: Assuming that the input sequence is camera motion aligned, the challenge is to merge them while considering the object motion. We begin by choosing an image with the least number of saturated pixels as the reference image. The final result will resemble the chosen reference image in the dynamic regions while combining HDR contents from all other images in static regions. Hence, different fusion rules are required depending on the presence or absence of object motion. Existing deep learning-based methods employ a monolithic CNN architecture to learn fusion rules applicable to both scenarios. Despite their success in typical scenes, they introduce artifacts in some challenging conditions with regions affected by both motion and saturation.

\begin{figure*}
\centering
  \includegraphics[width=18cm, clip, trim= 1.25cm 0.02cm 1.2cm 0.02cm]{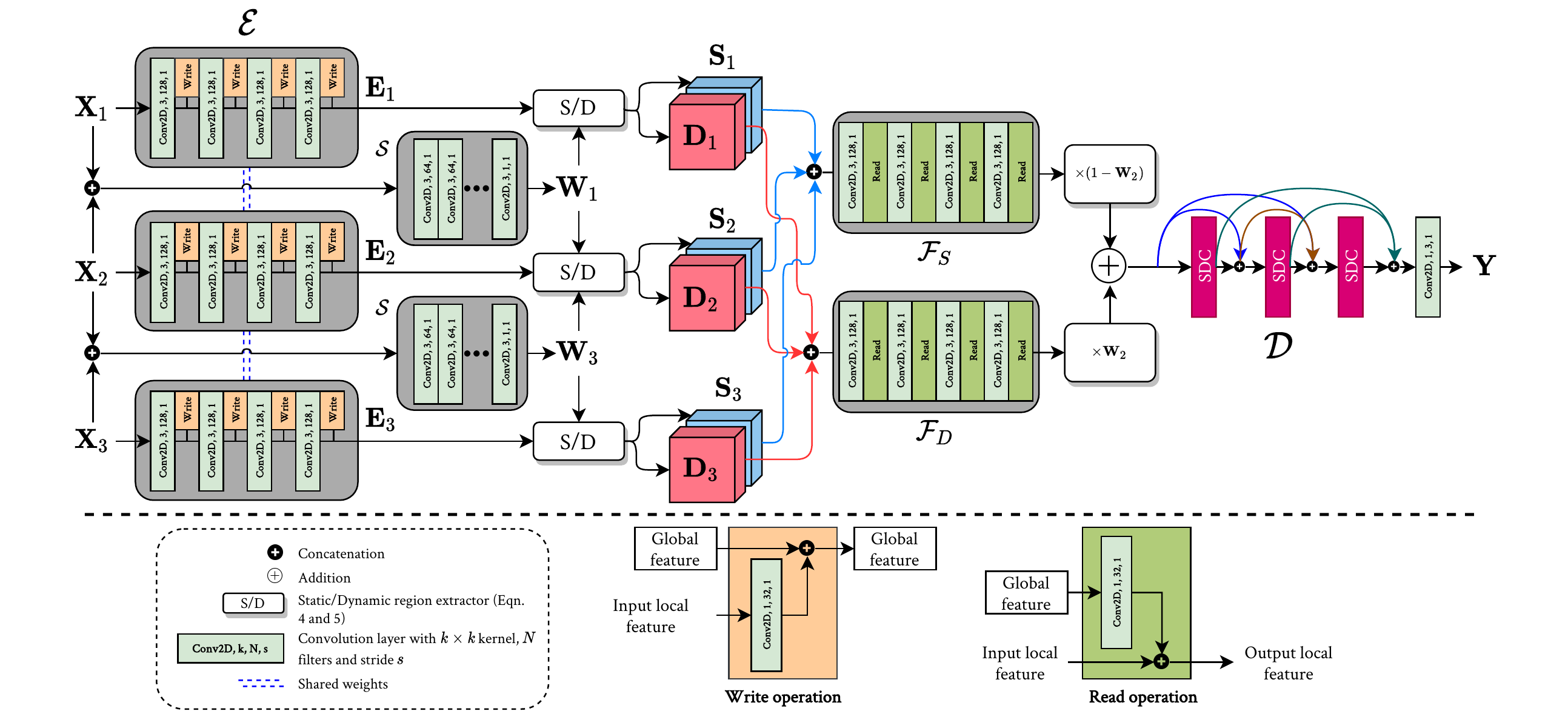}%
  \caption{Overview of our proposed approach. We use a simple four-layer encoder to generate $\mathbf{E}_i$ features for inputs $\mathbf{X}_i$. We then segment $\mathbf{E}_i$ into static ($\mathbf{S}_i$) and dynamic regions ($\mathbf{D}_i$) with the help of a segmentation map generated by $\mathcal{S}$. $\mathbf{S}_i$ and $\mathbf{D}_i$ of different input images are concatenated and fused by $\mathcal{F}_S$ and $\mathcal{F}_D$. Finally, the fused static and dynamic features are merged by a decoder to generate the final HDR image.} 
  \label{fig:prop_arch}
\end{figure*}

To address this problem, we segment the input images into static and dynamic regions (see Fig. \ref{fig:prop_arch}) and fuse them separately. Hence, the model has the freedom to learn different fusion rules for static and dynamic regions. Finally, we merge the fused static and dynamic features to generate fused HDR output.
    
\textbf{Formulation}: For ease of understanding, we explain our approach with three input varying exposure images, $N=3$; however, we can easily extend our method to fuse arbitrary-length sequence as explained in Section \ref{sec:discussion}. Following \cite{kalantari2017deep,wu2018deep,yan2019attention,prabhakar2019fast,prabhakar2020}, we choose the middle image ($\mathbf{X}_2$) as reference. The generated $\mathbf{Y}$ will have the same structural details as of reference in well-exposed regions of $\mathbf{X}_2$, and borrow details from other images in ill-exposed regions of $\mathbf{X}_2$.
\subsubsection{Feature encoder ($\mathcal{E}$)} We begin by extracting convolutional features for input varying exposure images using a encoder $\mathcal{E}$. As the HDR domain data of input LDR images help in locating misalignments, we concatenate both HDR and LDR domain of $\mathbf{X}_k$ to form a six-channel input to $\mathcal{E}$. The LDR images can be converted to HDR domain by, $\mathbf{H}_k=\mathbf{X}^{2.2}_k / t_k$, $\forall k = (1,2,3)$, where $t_k$ denote the exposure time of $\mathbf{X}_k$. Each input image is fed to a shared encoder to extract individual feature maps $\{\mathbf{E}_k\}$,
    \begin{equation}
        \mathbf{E}_k = \mathcal{E}\big(\{\mathbf{X}_k, \mathbf{H}_k\}\big), \forall k = (1,2,3), 
        \label{eqn:eqn1}
    \end{equation}
\subsubsection{Dynamic region segmentation ($\mathcal{S}$)} In order to identify dynamic regions with respect to the reference image, we use a separate segmentation network $\mathcal{S}$. We transfer the exposure of the reference image $\mathbf{X}_2$ to non-reference images $\mathbf{X}_1$ and $\mathbf{X}_3$:
\begin{equation}
    \mathbf{H^{ref}_i} = X^{2.2}_i/t_{ref}, \forall i = (1, 3)
\end{equation}

and concatenate non-reference image ($k=1,3$) with the reference image and pass as input to the segmentation network $\mathcal{S}$,
    \begin{equation}
        \mathbf{W}_i = \mathcal{S}\big( \{ \mathbf{H^{ref}_i}, \mathbf{H_{ref}} \} \big), \forall i = (1,3),
        \label{eqn:eqn2}
    \end{equation}
The output of $\mathcal{S}$ is a single channel segmentation map ($\mathbf{W}_i$) at the same resolution as input, with values in $\{0, 1\}$. While the value `0' indicates the absence of motion between $\mathbf{X}_i$ and $\mathbf{X}_2$, value `1' indicates the presence of motion. During this process, we generate one segmentation map for each one of the non-reference images. Then, we obtain the segmentation map for the reference image by taking a union of the non-reference image segmentation maps,
    \begin{equation}
        \mathbf{W}_2 = \mathbf{W}_1\displaystyle\bigcup\mathbf{W}_3,
        \label{eqn:eqn3}
    \end{equation}
$\mathbf{W}_2$ contains the motion in all sequences with respect to the reference image $\mathbf{X}_2$. Using the predicted segmentation maps $\mathbf{W}_k$, we divide the input image feature maps $\mathbf{E}_k$ into static ($\mathbf{S}_k$) and dynamic features ($\mathbf{D}_k$),
\begin{alignat}{2}
    \mathbf{D}_k &= \mathbf{W}_k \times \mathbf{E}_k, &\forall k = 1,2,3 \\
    \mathbf{S}_k &= \big[ 1 - \mathbf{W}_k \big] \times \mathbf{E}_k, &\forall k = 1,2,3
    \label{eqn:eqn4}
\end{alignat}
For $\mathcal{S}$, we use two separate U-net \cite{ronneberger2015u} styled architectures for predicting $\mathbf{W}_1$ and $\mathbf{W}_3$. It consists of an encoder with four convolution blocks, with max-pooling after each block. We pass the encoded features to a decoder with convolutional and upsampling layers. Additionally, we also concatenate the encoder features with the decoder using skip connections. 
\subsubsection{Fusion networks} As the static features do not have any moving regions, they can be fused by combining best-exposed image features across all images. For dynamic features, the structure should resemble the reference image in non-saturated regions and should have borrowed HDR details from other images in saturated regions. We achieve this objective by merging static and dynamic features using two separate convolutional blocks: $\mathcal{F}_S$ and $\mathcal{F}_D$, respectively.

The concatenated static features is fed as input to $\mathcal{F}_S$ and concatenated dynamic features to $\mathcal{F}_D$. As both $\mathcal{F}_S$ and $\mathcal{F}_D$ are separate convolutional blocks, they learn two different fusion rules for static and dynamic feature fusion. $\mathcal{F}_S$ and $\mathcal{F}_D$ outputs the fused static feature map ($\mathbf{S}_f$) and a fused dynamic feature map ($\mathbf{D}_f$),
        \begin{align}
            \mathbf{S}_f &= \mathcal{F}_S\bm{(} \mathrm{CONCAT}(\{\mathbf{S}_k\}) \bm{)}\\
            \mathbf{D}_f &= \mathcal{F}_D\bm{(} \mathrm{CONCAT}(\{\mathbf{D}_k\}) \bm{)}
            \label{eqn:eqn5}
        \end{align}
\subsubsection{Global Memory} The proposed approach's performance can be bolstered and generalized to $N$ images using a concept we call \textit{Global Memory}. The architecture features an external memory, which is essentially a collection of concatenated encoder feature maps. This memory is global in that it can be accessed using special convolutional blocks anywhere in the network. We use these blocks for writing to and reading from memory. The memory used here is unbounded; thus the size grows with the number of images. 

\textbf{Write operation}: The individual image features are added to the global memory using a write operation. Each write operation appends 32 dimension feature map to the memory (See Fig. \ref{fig:prop_arch}). The operation consists of two parts: convolution of the input features with trainable kernels and concatenation of the result to the global memory. We apply the write operation to the features extracted during the encoding process.  
    \LinesNotNumbered
    \begin{algorithm}[t]
    \label{alg:the_alg}
        \DontPrintSemicolon
          \KwIn{Varying exposure images $\{\mathbf{X}_k\}$ and their exposure times $t_k$, $\forall k = (1,2,3)$}
          \KwOut{Fused HDR image, $\mathbf{Y}$}
          \tcc{\small Extract features with Encoder($\mathcal{E}$)}
          \nextnr
          $\{\mathbf{E}_k\}$ = $\mathcal{E}\bm{(}\{\mathbf{X}_k, \mathbf{H}_k\}\bm{)}, where \: \mathbf{H}_k=\mathbf{X}^{2.2}_k / t_k$
          
          \tcc{\small Segment dynamic regions with $\mathcal{S}$}
          \nextnr
          $\{\mathbf{W}_i\}$ = $\mathcal{S}\bm{(}\{\mathbf{X}_i, \mathbf{X}_2\}\bm{)}, \forall i = (1,3)$ \\
          $\mathbf{W}_2$ = $\mathbf{W}_1\displaystyle\bigcup\mathbf{W}_3$
          
          \tcc{\small Extract dynamic and static parts}
          \nextnr
          $\{\mathbf{D}_k\}$ = $\{\mathbf{W}_k\} \times \{\mathbf{E}_k\}$\\
          $\{\mathbf{S}_k\}$ = $\bm{[} 1 - \{\mathbf{W}_k\} \bm{]} \times \{\mathbf{E}_k\}$
          
          \tcc{\small Fuse dynamic and static parts}
          \nextnr
          $\mathbf{D}_f$ = $\mathcal{F}_D\bm{(} \mathrm{CONCAT}(\{\mathbf{D}_k\}) \bm{)}$\\
          $\mathbf{S}_f$ = $\mathcal{F}_S\bm{(} \mathrm{CONCAT}(\{\mathbf{S}_k\}) \bm{)}$
          
          \tcc{\small Merge fused dynamic and static features}
          \nextnr
          $\mathbf{Z}$ = $\mathbf{W}_2 \times \mathbf{D}_f + \bm{(} 1 - \mathbf{W}_2 \bm{)} \times \mathbf{S}_f$
          
          \tcc{\small Generate fused HDRI with decoder($\mathcal{D}$)}
          \nextnr
          $\mathbf{Y}$ = $\mathcal{D}\bm{(} \mathbf{Z} \bm{)}$
        \caption{Overview of proposed method}
    \end{algorithm}

\textbf{Read operation}: The aggregated global memory features are tapped during the fusion process with a read operation. Each read from memory derives a fixed number of channels from memory. This number is also set to 32 channels in our implementation. The operation consists of a convolution of memory features with trainable kernels, followed by a concatenation of the results with the network's main branch. Since the number of channels in the memory is variable, we use grouped convolutions to reduce the channels to a fixed size. Each group's size depends on the number of channels in memory and the number of channels to be appended to the main network. Instead of concatenating the memory features during a read operation, we chose to distill the memory and extract only the necessary features required for the current stage. We achieve this by applying a 1$\times$1 convolution filter on the global memory. As shown in Fig. \ref{fig:prop_arch}, multiple individual read operations are applied during $\mathcal{F}_S$ and $\mathcal{F}_D$. 
\begin{figure}
    \centering
      \includegraphics[clip, trim=0.02cm 0.02cm 0.02cm 0.8cm, width=6cm]{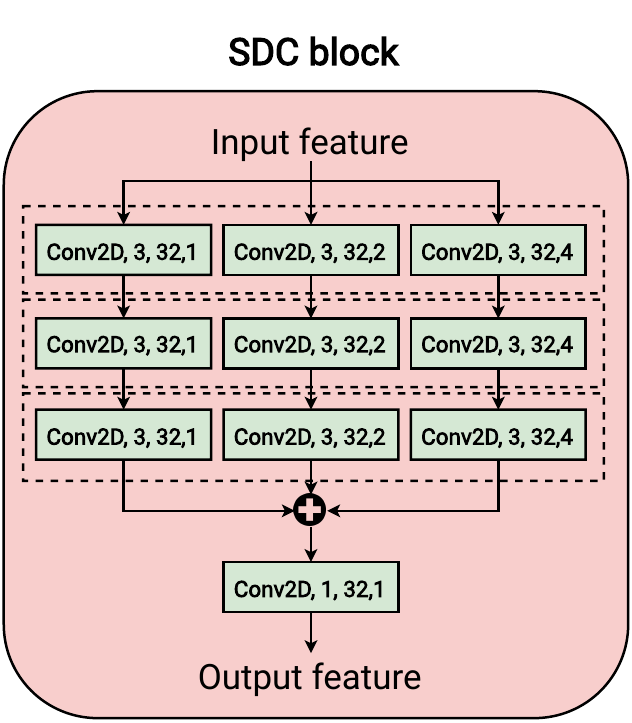}%
      \caption{Architecture of SDC block.}
      \label{fig:prop_sdc_block}
    \end{figure}
\subsubsection{Feature decoder ($\mathcal{D}$)} The fused static ($\mathbf{S}_f$) and dynamic ($\mathbf{D}_f$) features are combined using the reference image segmentation map, $\mathbf{W}_2$. The non-zero pixels in $\mathbf{W}_2$ indicate the dynamic regions in all input images, and zero pixels indicate otherwise. Hence, $\mathbf{D}_f$ is chosen in the dynamic regions and $\mathbf{S}_f$ in the static regions by,
\begin{equation}
    \mathbf{Z} = \mathbf{W}_2 \times \mathbf{D}_f + \bm{(} 1 - \mathbf{W}_2 \bm{)} \times \mathbf{S}_f
\end{equation}
We pass the combined feature $\mathbf{Z}$ as input to a series of three densely connected Stacked and Dilated Convolutional (SDC) blocks. SDC blocks introduced by Schuster \textit{et al.} \cite{schuster2019sdc} consists of four parallel convolutional layers with dilation rate of $(1,2,3,4)$. We modify the original SDC block to have three stacked convolutional layers with dilation rate of $(1,2,4)$ (see Fig. \ref{fig:prop_sdc_block}). Additionally, to reduce the number of parameters, the stacked convolutional layers share the weights between them. An SDC block consists of three concurrent stacked convolutional layers. The stacked convolutional layers contain 32 filters with 3$\times$3 kernel. The last three stacked layers' features are concatenated and further processed by a 1$\times$1 convolution layer.

The individual outputs of each SDC block are concatenated and passed to a 1$\times$1 convolutional layer to generate the final output $\mathbf{Y}$. 
\subsubsection{Loss functions} The loss to train our model consists of two sub losses. The first sub-loss is computed between the predicted ($\mathbf{W}_i$) and ground truth ($\hat{\mathbf{W}}_i$) segmentation maps as,
\begin{equation}
    \mathcal{L}_{seg} = E_{D}(\mathbf{W}_1, \hat{\mathbf{W}}_1) + E_{D}(\mathbf{W}_3, \hat{\mathbf{W}}_3)
\end{equation}
where $E_{D}$ denotes the Dice loss error function (\nolinebreak\hspace{1sp}\cite{deng2018learning}). The second sub-loss consists of $\ell_2$ loss between predicted $\mathbf{Y}$ and ground truth HDR ($\hat{\mathbf{Y}}$) images. As it is conventional to display the HDR images after tone mapping, we compute the loss after tone mapping the HDR images with $\mu$-law operator. The $\mu$-law tone mapping function is defined as,
\begin{equation}
    T(\mathbf{Y}) = \dfrac{log(1+\mu \mathbf{Y})}{log(1+\mu)}
    \label{eqn:mu_law}
\end{equation}
where $\mu$ = 5000. Then, the second sub-loss is defined as,
\begin{equation}
    \mathcal{L}_{HDR} = \ell_2(T(\mathbf{Y}), T(\hat{\mathbf{Y}}))
\end{equation}
where $\ell_2$ denotes the mean squared error function. The final loss is defined as weighted sum of two sub-losses,
\begin{equation}
    \mathcal{L}_{final} = \alpha \times \mathcal{L}_{seg} + \beta \times \mathcal{L}_{HDR}
\end{equation}
To ensure convergence, we initially use high $\alpha$ of 1 with low $\beta$ value of $10^{-4}$. Then after 50 epochs, we change $\alpha$ to $10^{-2}$ and $\beta$ to 1.
\section{Evaluation and Results}
    \label{sec:eval}
    \subsection{Multi-Exposure Dynamic Motion Segmentation dataset} In the field of HDR deghosting, since most approaches involve non-deep methods for solving the problem, there is a lack of a large data repository of multi-exposure image sequences needed for training deep neural networks. To tackle this issue, we create MEDS - a Multi-Exposure Dynamic motion Segmentation dataset. MEDS consists of 3683 multi-exposure images with segmentation annotations of moving regions.  
    \begin{figure*}[ht]
    \centering
      \includegraphics[width=17cm]{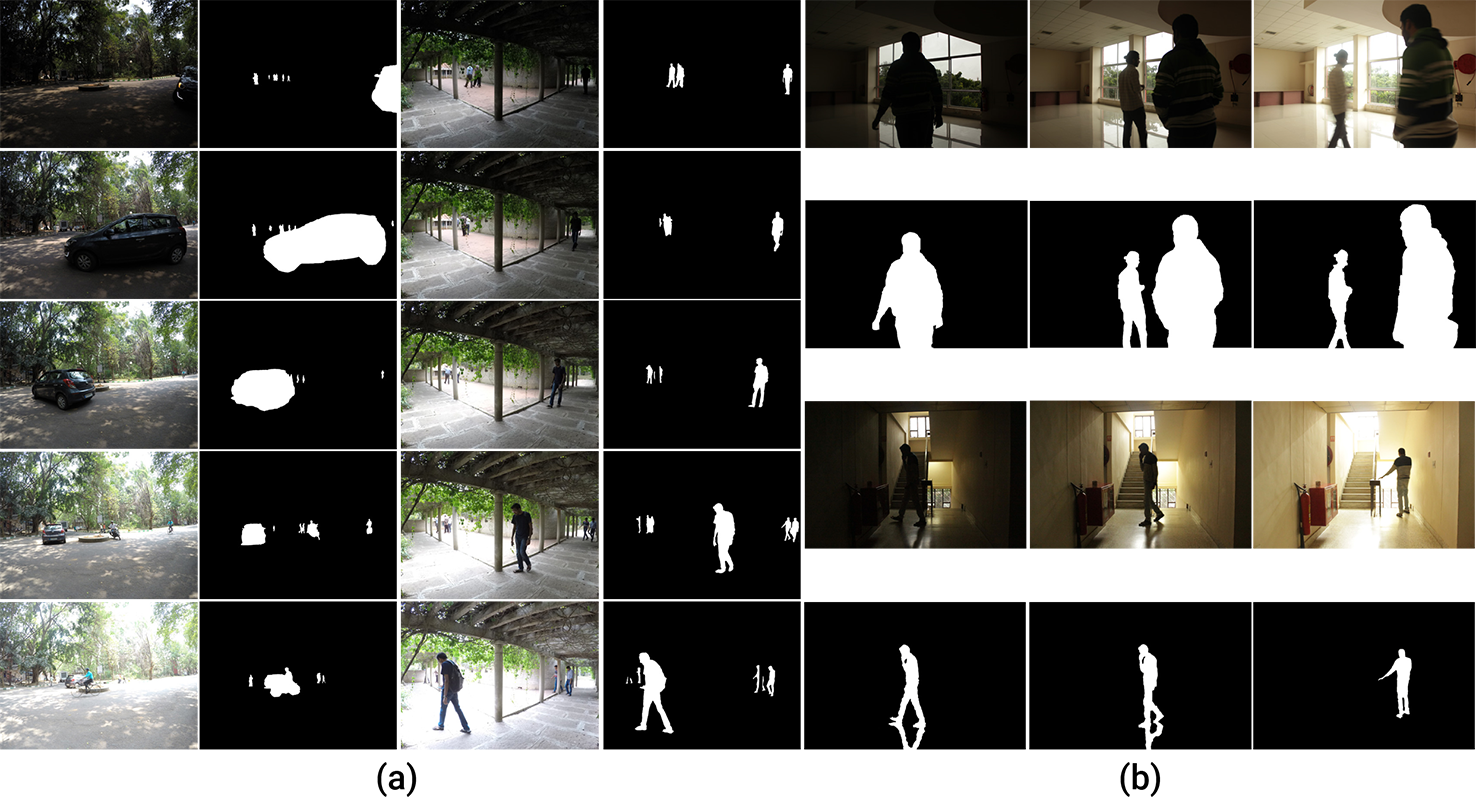}%
      \caption{Example images and their annotated motion maps from MEDS dataset. (a) Images captured with GoPro Hero 5 camera, (b) Images captured with Canon EOS 500D and 600D cameras.}
      \label{fig:meds}
    \end{figure*}
Among the 3683 images, 2660 images were captured with a GoPro Hero 5 camera. Each exposure stack in this set has five images with $\{-2, -1, 0, +1, +2\}$EV bias. The remaining 1023 images were captured using Canon EOS 500D and 600D cameras with each exposure stack having sequences of three images with $\{-2, 0, +2\}$EV bias. The images were captured with a tripod-mounted camera in Auto-exposure bracketing mode by varying only the exposure time while the ISO and aperture values are fixed. The dataset captures a wide variety of scenes (indoor, outdoor, landscapes), different lighting conditions with varying natural motion (slow-moving persons to fast-moving vehicles). Sample images from the MEDS dataset are shown in Fig. \ref{fig:meds}.

\textbf{Annotation}: For addressing the task of dynamic moving object segmentation from the registered input images, we provide human-annotated motion masks for 3683 scenes. The annotators visually inspect a GIF image created from the sequence for a few seconds to identify the dynamic regions. Then, the annotators draw a boundary around the moving regions. For each image, we obtain two different annotations by two different individuals. The final annotation for the image is obtained by taking the union of the two. In mismatch regions between the two, a final annotator inspects the difference and chooses the correct one. In total, six annotators were involved in creating the segmentation dataset. We plan to release this dataset of images and their corresponding dynamic moving object annotations publicly to benefit the HDR community. In Fig. \ref{fig:result3}, we present example images, their ground truth, and predicted segmentation maps.
\subsection{Implementation} We have implemented our model in TensorFlow (\nolinebreak\hspace{1sp}\cite{abadi2016tensorflow}). The weights are initialized with Glorot uniform initialization (\nolinebreak\hspace{1sp}\cite{glorot2010understanding}) and was trained on a machine with Intel core i7-9700F CPU and a NVIDIA Quadro RTX 6000 GPU. We use Adam optimizer (\nolinebreak\hspace{1sp}\cite{kingma2014adam}) with $10^{-4}$ learning rate and a batch size of four to train the model for 200 epochs. The learning rate is reduced by a factor of 0.96 after every epoch. We train the model using Kalantari \textit{et al.} \cite{kalantari2017deep} and Prabhakar \textit{et al.} \cite{prabhakar2019fast} datasets. \cite{kalantari2017deep} dataset consists of 74 training and 15 validation sequences, while \cite{prabhakar2019fast} dataset consists of 466 training and 116 validation sequences. In addition to \cite{kalantari2017deep} and \cite{prabhakar2019fast}, we have also validated our model on publicly available datasets like Sen \textit{et al.} \cite{sen2012robust} and Tursun \textit{et al.} \cite{tursun2016objective}.

As the ground truth deghosted HDR image is unavailable for MEDS dataset images, they were used to pre-train the segmentation network only. Later, the pre-trained segmentation network weights are used as initialization while training with \cite{kalantari2017deep} dataset. We train the model with random patches of size 128$\times$128 cropped from random locations in the full image. 
\subsection{Quantitative results}
\label{subsec:quant_results}
    \begin{table}[t]
        \caption{Quantitative comparison between proposed method against nine state-of-the-art methods. Best score highlighted in Blue with bold and second-best score is highlighted in gray. Refer Section \ref{subsec:quant_results} for more details.}
        \setlength{\tabcolsep}{2pt}
        \centering
        \subfloat[][Quantitative evaluation on \cite{kalantari2017deep} dataset.]{
        \label{tab:ucsd_eval}
        \begin{tabular}{@{}lccccccccccc@{}}
        \toprule
        \multicolumn{1}{c}{} & \multicolumn{5}{c}{PSNR} &  \multicolumn{5}{c}{SSIM} &  \\ \cmidrule(lr){2-6} \cmidrule(lr){7-11}
        \multicolumn{1}{c}{} & T1 & T2 & T3 & T4 & T5 & T1 & T2 & T3 & T4 & T5 & \multirow{-2}{*}{\begin{tabular}[c]{@{}c@{}}HDR-\\ VDP-2\end{tabular}} \\ \midrule
        M1        & 31.25 & 35.75 & 30.18 & 28.87 & 31.69 &  0.941 & 0.963 & 0.940 & 0.942 & 0.944 & 62.07 \\
        M2       & 38.57 & 40.94 & 27.81 & 30.33 & 31.35 &  0.971 & 0.978 & 0.955 & 0.966 & 0.969 & 64.74 \\
        M3         & 8.846 & 21.33 & 12.26 & 12.10 & 20.65 &  0.107 & 0.622 & 0.715 & 0.787 & 0.738 & 54.00 \\
        M4      & 14.21 & 14.13 & 25.23 & 20.26 & 26.82 & 0.350 & 0.882 & 0.925 & 0.923 & 0.935 & 57.95 \\
        M5 & 41.27 & \cellcolor[HTML]{C0C0C0}42.74 & 34.12 & 33.70 & 32.99 & 0.981 & 0.987 & 0.980 & 0.979 & 0.974 & 66.10 \\
        M6        & 40.91 & 41.65 & 34.98 & 34.54 & 33.69 &   0.986 & 0.986 & 0.982 & 0.982 & 0.977 & \cellcolor[HTML]{C0C0C0}67.44 \\
        M7 & 39.68 & 40.47 & 33.56 & 34.08 & 32.60 &  0.980 & 0.975 & 0.966 & 0.975 & 0.961 & 66.50 \\
        M8 & \cellcolor[HTML]{C0C0C0}41.33 & \cellcolor[HTML]{A0A2ED}\textbf{42.82} & \cellcolor[HTML]{C0C0C0}35.18 & \cellcolor[HTML]{C0C0C0}36.94 & \cellcolor[HTML]{C0C0C0}36.29 & 0.986 & \cellcolor[HTML]{C0C0C0}0.989 & \cellcolor[HTML]{C0C0C0}0.984 & \cellcolor[HTML]{C0C0C0}0.985 & \cellcolor[HTML]{C0C0C0}0.982 & 67.15 \\
        M9       & 41.08 & 41.21 & 29.83 & 33.28 & 29.51 &  \cellcolor[HTML]{C0C0C0}0.989 & \cellcolor[HTML]{C0C0C0}0.989 & 0.962 & 0.978 & 0.974 & \cellcolor[HTML]{A0A2ED}\textbf{67.53} \\
        M10    & \cellcolor[HTML]{A0A2ED}\textbf{41.47} & 42.03 & \cellcolor[HTML]{A0A2ED}\textbf{35.70} & \cellcolor[HTML]{A0A2ED}\textbf{38.48} & \cellcolor[HTML]{A0A2ED}\textbf{36.67} &  \cellcolor[HTML]{A0A2ED}\textbf{0.992} & \cellcolor[HTML]{A0A2ED}\textbf{0.992} & \cellcolor[HTML]{A0A2ED}\textbf{0.988} & \cellcolor[HTML]{A0A2ED}\textbf{0.989} & \cellcolor[HTML]{A0A2ED}\textbf{0.986} & 67.26 \\ \bottomrule
        \end{tabular}}
        \\ 
        \subfloat[][Quantitative evaluation on \cite{prabhakar2019fast} dataset. The abbrevations used are as follows: T1 - linear domain, T2 - $\mu$-law tonemapper, T3 - Krawczyk \textit{et al.} \cite{krawczyk2005lightness}, T4 - Reinhard \textit{et al.} \cite{reinhard2005dynamic}, and T5 - Durand \textit{et al.} \cite{durand2002fast} tonemappers. M1 - Hu13, M2 - Sen12, M3 - Endo17, M4 - Eilertsen17, M5 - Kalantari17, M6 - Wu18, M7 - Prabhakar19, M8 - Prabhakar20, M9 - Yan19, and M10 - Proposed method.]{
        \label{tab:iccp_eval}
        \begin{tabular}{@{}lccccccccccc@{}}
        \toprule
        \multicolumn{1}{c}{} & \multicolumn{5}{c}{PSNR} &  \multicolumn{5}{c}{SSIM} & \\ \cmidrule(lr){2-6} \cmidrule(lr){7-11}
        \multicolumn{1}{c}{} & T1 & T2 & T3 & T4 & T5 &  T1 & T2 & T3 & T4 & T5 & \multirow{-2}{*}{\begin{tabular}[c]{@{}c@{}}HDR-\\ VDP-2\end{tabular}} \\ \midrule
        M1        & 29.47 & 32.58 & 27.61 & 26.84 & 26.20 &   0.954 & 0.949 & 0.912 & 0.919 & 0.917 & 63.50 \\
        M2       & 32.93 & 33.43 & 30.22 & 29.15 & 30.91 &   0.972 & 0.964 & 0.950 & 0.948 & 0.950 & 65.47 \\
        M3      & 9.760 & 8.980 & 11.18 & 13.07 & 20.60 &   0.132 & 0.641 & 0.675 & 0.763 & 0.711 & 55.76 \\
        M4 & 14.19 & 15.66 & 22.47 & 22.04 & 24.97 &   0.442 & 0.869 & 0.879 & 0.897 & 0.904 & 58.74 \\
        M5 & 32.50 & 35.63 & 30.08 & 28.81 & 30.45 &   0.969 & 0.961 & 0.938 & 0.943 & 0.940 & 65.40 \\
        M6        & 34.40 & 38.03 & 33.35 & 31.82 & 32.66 &  0.977 & 0.971 & 0.962 & 0.957 & 0.956 & 66.59 \\
        M7 & 32.74 & 36.08 & 30.66 & 29.83 & 30.54 &  0.967 & 0.959 & 0.942 & 0.935 & 0.939 & 66.10 \\
        M8 & 34.98 & 38.30 & 32.99 & 31.76 & 32.53 &  0.978 & 0.970 & 0.960 & 0.952 & 0.953 & 66.25 \\
        M9 & \cellcolor[HTML]{C0C0C0}35.28 & \cellcolor[HTML]{C0C0C0}38.65 & \cellcolor[HTML]{C0C0C0}33.82 & \cellcolor[HTML]{C0C0C0}32.08 & \cellcolor[HTML]{A0A2ED}\textbf{33.31} &  \cellcolor[HTML]{C0C0C0}0.980 & \cellcolor[HTML]{C0C0C0}0.973 & \cellcolor[HTML]{C0C0C0}0.963 & \cellcolor[HTML]{C0C0C0}0.961 & \cellcolor[HTML]{C0C0C0}0.957 & \cellcolor[HTML]{C0C0C0}66.88 \\
        M10    & \cellcolor[HTML]{A0A2ED}\textbf{37.21} & \cellcolor[HTML]{A0A2ED}\textbf{39.71} & \cellcolor[HTML]{A0A2ED}\textbf{34.99} & \cellcolor[HTML]{A0A2ED}\textbf{32.83} & \cellcolor[HTML]{C0C0C0}33.25 &   \cellcolor[HTML]{A0A2ED}\textbf{0.984} & \cellcolor[HTML]{A0A2ED}\textbf{0.977} & \cellcolor[HTML]{A0A2ED}\textbf{0.969} & \cellcolor[HTML]{A0A2ED}\textbf{0.965} & \cellcolor[HTML]{A0A2ED}\textbf{0.962} & \cellcolor[HTML]{A0A2ED}\textbf{68.03} \\ \bottomrule
        \end{tabular}}
        \label{tab:main_eval}
    \end{table}
    In Table \ref{tab:main_eval}, we report the quantitative results comparison between proposed method and nine state-of-the-art approaches for \cite{kalantari2017deep} and \cite{prabhakar2019fast} datasets. The compared methods are: \begin{enumerate*}[label=(\roman*)]
      \item Hu13 - Hu \textit{et al.} \cite{hu2013hdr},
      \item Sen12 - Sen \textit{et al.} \cite{sen2012robust},
      \item Endo17 - Endo \textit{et al.} \cite{endo2017deep},
      \item Eilertsen17 - Eilertsen \textit{et al.} \cite{eilertsen2017hdr},
      \item Kalantari17 - Kalantari \textit{et al.} \cite{kalantari2017deep},
      \item Wu18 - Wu \textit{et al.} \cite{wu2018deep},
      \item Prabhakar19 - Prabhakar \textit{et al.} \cite{prabhakar2019fast},
      \item Prabhakar20 - Prabhakar \textit{et al.} \cite{prabhakar2020}
      \item Yan19 - Yan \textit{et al.} \cite{yan2019attention}.
    \end{enumerate*}
    Among the nine comparison methods, Hu13 and Sen12 are classical non-deep approaches, whereas Endo17 and Eilertsen17 are deep learning-based single image HDR generation methods. The rest of them are deep learning-based state-of-the-art approaches for HDR deghosting.
    
    For evaluation, we use three full-reference metrics: PSNR, SSIM \cite{wang2004image}, and HDR-VDP-2 \cite{mantiuk2011hdr}. HDR-VDP-2 is a full reference metric designed specifically to evaluate HDR images in the linear domain. We compute PSNR and SSIM in the linear domain and after applying four different tone mapping operators. One of them is the $\mu$-law tonemapper, which was used to train the network in Eqn. \ref{eqn:mu_law}. Hence, we evaluate with three different tone mappers other than the $\mu$-law tone mapper for more representative or realistic evaluation. Those three tone mappers are: Krawczyk \textit{et al.} \cite{krawczyk2005lightness}, Reinhard \textit{et al.} \cite{reinhard2002photographic} and Durand \textit{et al.} \cite{durand2002fast}. It should be noted that the network was not trained to optimize these three tone mappers. In total, we compare our proposed approach against nine state-of-the-art methods using eleven metrics. 
    \begin{table}[t]
    \centering
    \caption{Ablation experiments. Refer Section \ref{subsec:ablations} for more details.}
    \label{tab:ablation_expts}
    \begin{tabular}{@{}llcc@{}}
    \toprule
    Method & \multicolumn{1}{c}{Description} & PSNR-L & PSNR-T \\ \midrule
    A1 & End-to-end training & 41.07 & 41.48 \\
    A2 & End-to-End training + segmentation loss & 40.87 & 41.38 \\
    A3 & No segmentation map & 40.58 & 40.69 \\
    A4 & Simple difference as segmentation map & 40.37 & 41.13 \\
    A5 & Simple difference + Triangle fusion & 34.40 & 33.50 \\
    A6 & CNN segmentation + Triangle fusion & 36.75 & 34.30 \\
    A7 & Arbitrary size memory network & 41.05 & 41.91 \\
    A8 & Single R/W submodule & 40.54 & 40.63 \\ \midrule
    \multicolumn{4}{c}{Decoder architecture ablations} \\ \midrule
    A9 & Vanilla & 41.14 & 41.31 \\
    A10 & Res-Net \cite{he2016deep} & 41.11 & 41.95 \\
    A11 & SDC without dense connection & 41.36 & 42.02 \\
    A12 & SDC with dense connection & \cellcolor[HTML]{A0A2ED}\textbf{41.47} & \cellcolor[HTML]{A0A2ED}\textbf{42.03} \\ \bottomrule
    \end{tabular}
    \end{table}
    
    In Table \ref{tab:main_eval}a, we present the quantitative comparison on Kalantari \textit{et al.} \cite{kalantari2017deep} dataset. In terms of PSNR, the proposed method outperforms others in all tone mappers except $\mu$-law tone mapper. In terms of SSIM, the proposed method outperforms others in all categories. In Table \ref{tab:main_eval}b, we present the quantitative comparison on Prabhakar \textit{et al.} \cite{prabhakar2019fast} dataset. The proposed method outperforms all other approaches in ten out of eleven metrics.
    \begin{figure*}[ht]
    \centering
      \includegraphics[width=16cm]{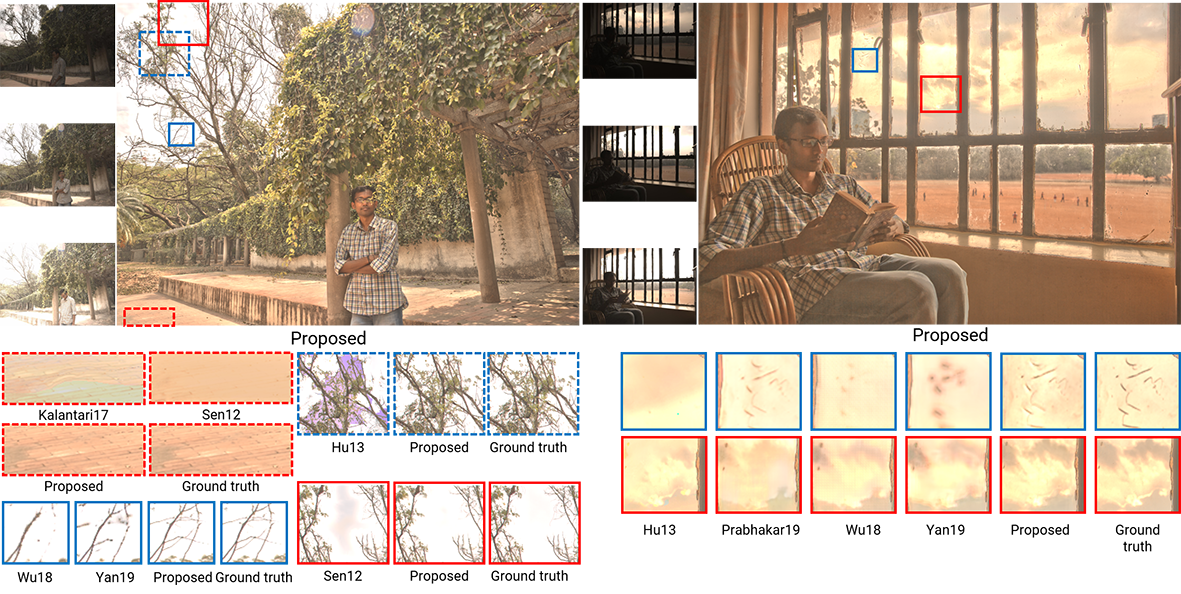}%
      \caption{Qualitative comparison between different methods for examples from \cite{prabhakar2019fast} dataset. The results are shown after tonemapping with Durand \textit{et al.} \cite{durand2002fast} method. Image best viewed zoomed in electronic monitors.}
      \label{fig:result2}
    \end{figure*}
\subsection{Ablation analysis}
    \label{subsec:ablations}
\begin{figure}[ht]
    \centering
      \includegraphics[width=8cm]{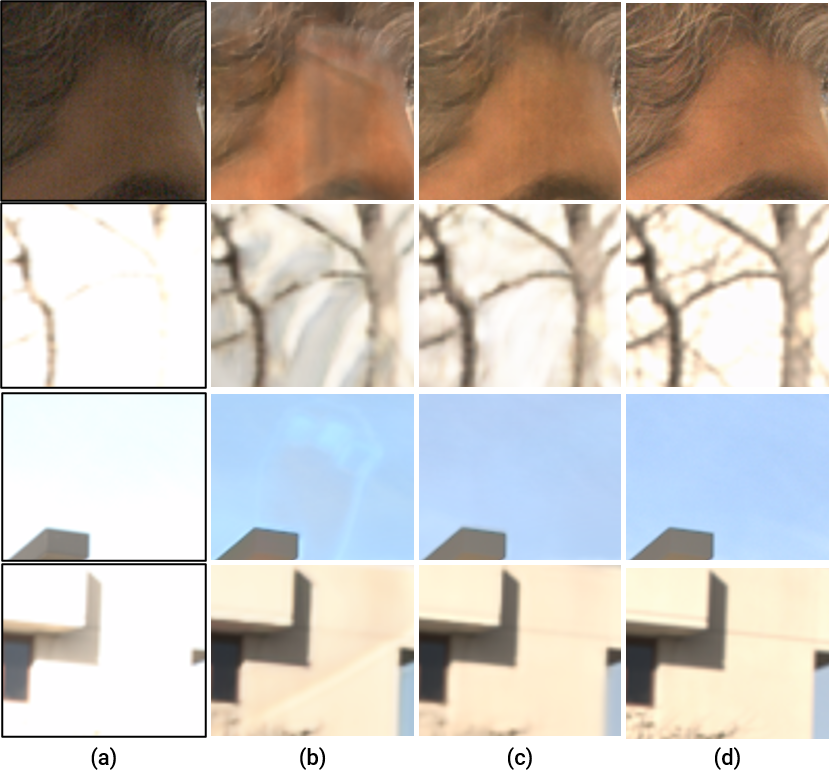}%
      \caption{Qualitative results to highlight the importance of segmentation maps. (a) Cropped and zoomed region from the reference image, (b) Result by the A3 model trained without segmentation maps, (c) Result by the proposed model using segmentation maps, and (d) Ground truth. A3 model suffers from mild ghosting artifacts prominent around the edges of moving regions.}
      \label{fig:ablation1}
    \end{figure}

\begin{figure}[th]
    \centering
      \includegraphics[width=8cm]{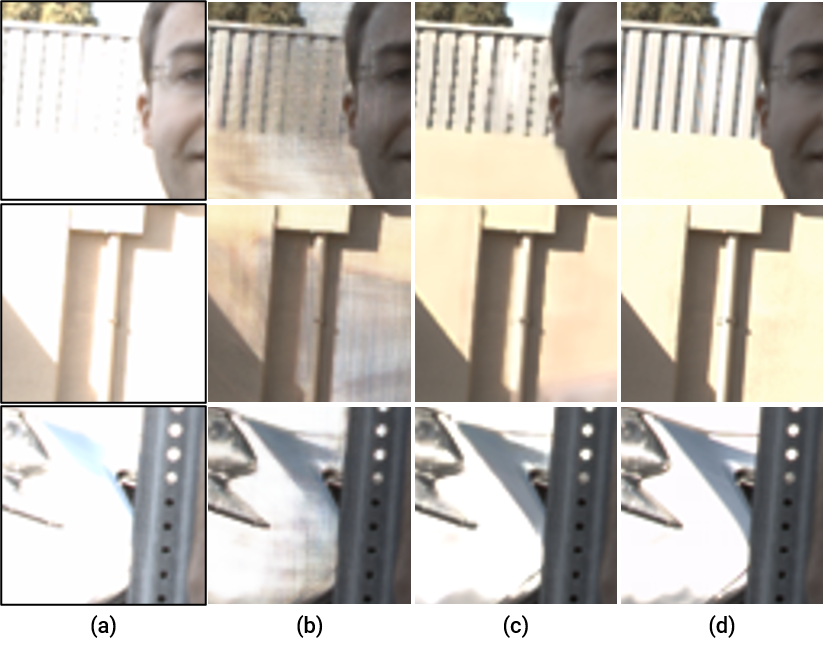}%
      \caption{Qualitative results to highlight the importance of memory network. (a) Cropped and zoomed region from the reference image, (b) Result by the model trained without memory network, (c) Result by the proposed model using memory network, and (d) Ground truth. As highlighted, using memory network aides the network to generate plausible textures in regions affected by saturation and motion.}
      \label{fig:ablation2}
    \end{figure}

    In Table \ref{tab:ablation_expts}, we present results for various baseline ablation experiments. All these ablation experiments were trained and tested on Kalantari \textit{et al.} \cite{kalantari2017deep} dataset. 
    \begin{table}[t]
        \caption{Comparison of two models trained with and without memory network for shared and unshared weights between $\mathcal{F}_S$ and $\mathcal{F}_D$. Refer Section \ref{subsec:ablations} for more details.}
        \label{tab:ablations_memory}
        \setlength{\tabcolsep}{2pt}
        \centering
        \begin{tabular}{@{}lccccc@{}}
        \toprule
         & \multicolumn{2}{c}{PSNR-L} &  & \multicolumn{2}{c}{HDR-VDP-2} \\ \cmidrule(lr){2-3} \cmidrule(l){5-6} 
         & Shared & Unshared &  & Shared & Unshared \\ \midrule
        With memory & 41.10 & \cellcolor[HTML]{A0A2ED}\textbf{41.47} &  & 67.11 & \cellcolor[HTML]{A0A2ED}\textbf{67.26} \\
        Without memory & 40.54 & 40.68 &  & 67.01 & 66.90 \\ \bottomrule
        \end{tabular}
    \end{table}
    \begin{itemize}[itemindent=1em,noitemsep,topsep=0pt,itemsep=0pt]
      \item[(A1)] We train our proposed method in an end-to-end fashion. Similar to Yan \textit{et al.} \cite{yan2020deep}, we don't enforce loss for segmentation network. The final HDR reconstruction loss is used to update both segmentation as well as fusion network. 
      \item[(A2)] With the same setting as A1, we train the model to minimize both HDR reconstruction and the segmentation loss. This baseline experiment performs slightly poorer than A1. This is because the fusion network is dependent on the segmentation map for its operation. However, by updating both segmentation and fusion network, the training becomes unstable and stuck at local minima.
      \item[(A3)] The segmentation map is not used to divide the features into dynamic and static features. $\{\mathbf{E}_k\}$ features are fused by a single fusion network and further processed by SDC blocks to generate the final result. As shown in Fig. \ref{fig:ablation1}, without the segmentation maps, the output still has some mild ghosting artifact around the edges.
      \item[(A4)] Instead of using a CNN to predict the segmentation map, we threshold the difference between brightness normalized source and reference image with 0.1 value to generate the motion map. Such a motion segmentation map has artifacts; even the illumination changes are detected as dynamic regions. Due to this, the overall PSNR is lower than predicting the motion map with a CNN. 
      \item[(A5)] The motion map generated by the simple difference approach from A4 is used to compensate for motion among the input images. For the dynamic regions, we replace source image pixels with exposure compensated reference image pixels. This results in three static varying exposure images. Then, the static images are fused using standard triangle function \cite{debevec2008recovering}. As expected, the segmentation error propagates to the final result without any correction. Hence, performance is inferior.
      \item[(A6)] We replace the simple difference map in A5 with the motion map predicted by the trained CNN network. This method performs better than A5; however, the CNN model still produces output with few artifacts. 
      \item[(A7)] We extend the proposed approach to fuse an arbitrary number of input images. See Section \ref{sec:discussion} for more details.  
      \item[(A8)] Instead of using different convolution weights for each instance of reading and writing operation, we share the weights across all read operations. Similarly, we share the weights across all write operations.
      \item[(A9)] We use nine vanilla convolution layers instead of dense SDC blocks.
      \item[(A10)] We use Res-Net \cite{he2016deep} for decoder model. 
      \item[(A11)] SDC without dense connection is used.
      \item[(A12)] SDC with dense connection is used for decoder architecture.
    \end{itemize}
    In Table \ref{tab:ablations_memory}, we present ablations results for using memory and sharing fusion networks. By sharing the fusion networks $\mathcal{F}_S$ and $\mathcal{F}_D$, a single fusion rule is used to merge both static and dynamic regions. As evidenced by both PSNR-L and HDR-VDP-2 metrics, having a memory network with unshared fusion networks outperforms other ablation methods. In Fig. \ref{fig:ablation2}, we present qualitative results generated by models trained with and without memory features. As seen in the figure, the network trained without memory fails to hallucinate viable details in saturated and occluded regions. With the support of memory features, the model can choose details for filling in saturated regions and avoid unnatural artifacts.
    \begin{figure*}[ht]
    \centering
      \includegraphics[width=17cm]{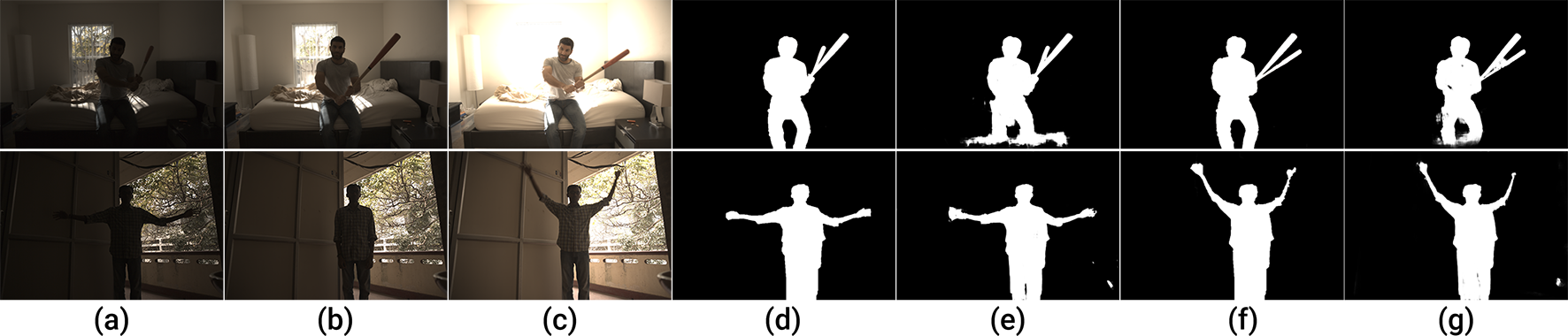}%
      \caption{(a)-(c) Input sequence, (d) - combined ground-truth motion segmentation map of (a) and (b), (e) - predicted motion segmentation map for (a) and (b), (f) - combined ground-truth motion segmentation map of (b) and (c), and (g) - predicted motion segmentation map of (b) and (c). The segmentation model has learned to differentiate between motion and saturation, and accurately segment only the moving regions with precise boundaries.}
      \label{fig:result3}
    \end{figure*}
    In Table \ref{tab:ablations_loss}, we show ablation results with using combinations of different loss functions on the tonemapped HDR images.
    \begin{table}[t]
    \centering
    \caption{Loss function ablations. Refer Section \ref{subsec:ablations} for more details.}
    \label{tab:ablations_loss}
    \begin{tabular}{@{}
    >{\columncolor[HTML]{FFFFFF}}l 
    >{\columncolor[HTML]{FFFFFF}}c 
    >{\columncolor[HTML]{FFFFFF}}c @{}}
    \toprule
    Methods & PSNR-L                                   & PSNR-T                                   \\ \midrule
    $\ell_2$  & \cellcolor[HTML]{A0A2ED}\textbf{41.47} & \cellcolor[HTML]{A0A2ED}\textbf{42.03} \\
    $\ell_1$  & 41.01   & 41.25  \\
    $\ell_2$ + $\ell_1$  & 40.85 & 41.32             \\
    $\ell_1$ + MS-SSIM  & \cellcolor[HTML]{C0C0C0}41.07 & \cellcolor[HTML]{C0C0C0}41.40 \\
    $\ell_2$ + MS-SSIM  & 40.13 & 39.83 \\
    $\ell_1$ + $\ell_2$ + MS-SSIM   & 40.61 & 40.63 \\ \bottomrule
    \end{tabular}
    \end{table}
    \begin{figure}[ht]
    \centering
      \includegraphics[width=8.5cm]{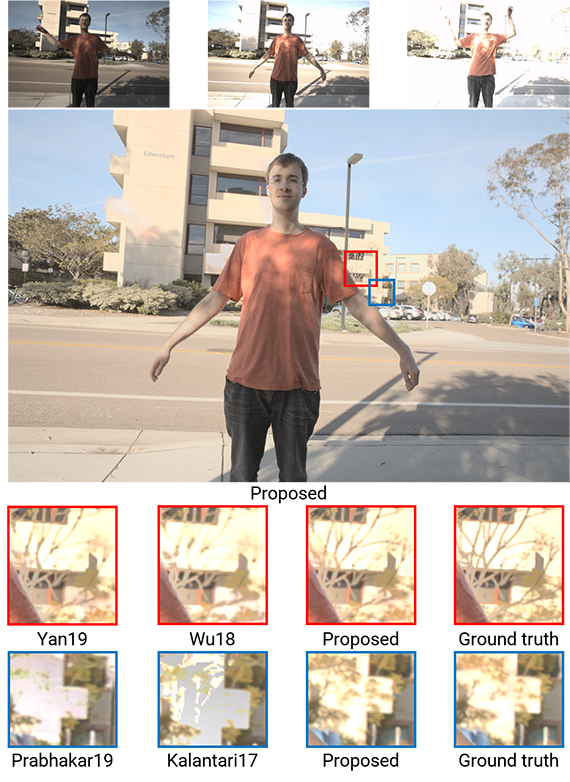}%
      \caption{Qualitative comparison between proposed and state-of-the art methods for a hard validation example from \cite{kalantari2017deep} dataset.}
      \label{fig:result4}
    \end{figure}
    \begin{figure}[ht]
    \centering
      \includegraphics[width=8.5cm]{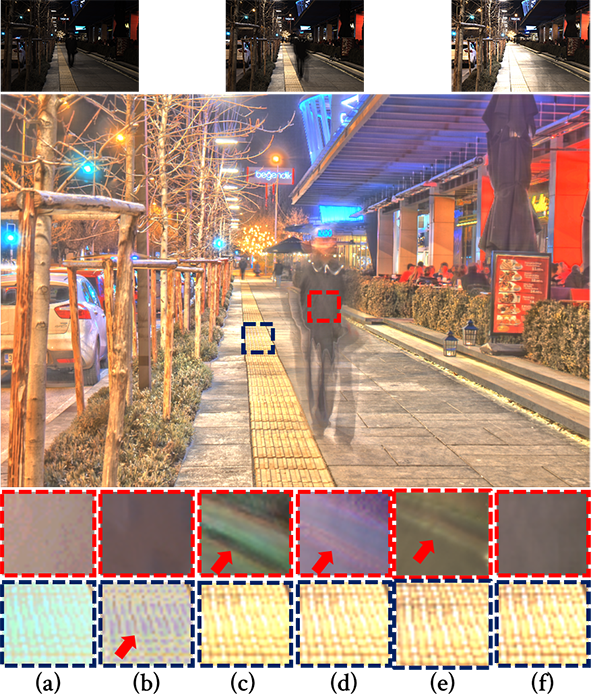}%
      \caption{Qualitative results by different methods for an example from METU dataset \cite{tursun2016objective}. (a) Hu13 \cite{hu2013hdr}, (b) Sen12 \cite{sen2012robust}, (c) Prabhakar19 \cite{prabhakar2019fast}, (d) Yan19 \cite{yan2019attention}, (e) Prabhakar20 \cite{prabhakar2020}, and (f) Proposed method. The highlighted zoomed regions show that existing methods have artifacts in both moving and saturated regions. Comparatively, our proposed method generates a ghosting-free result.}
      \label{fig:result5}
    \end{figure}
\subsection{Qualitative results}
    In Fig. \ref{fig:result1}, we compare proposed method against state-of-the-art HDR deghosting methods for a difficult validation image from \cite{kalantari2017deep} dataset\footnote{Please refer to supplementary material for additional results.}. It is a challenging sequence, as the moving arm occludes the reference image saturated regions in the low exposure image. The traditional methods such as Sen12 and Hu13 were unable to hallucinate correct textures in those regions. In those regions, the reference image does not possess any concrete structural information, and additionally, those regions are occluded by the moving arm in the low exposure image. Thus, the patch-match based synthesis methods fail to find correct details and fill those regions with false-color details (black zoom box in Fig. \ref{fig:result1}). Comparatively, our proposed method can produce textures with plausible details, even in such challenging regions. In Fig. \ref{fig:result2}, Sen12 method exhibits similar false color artifacts in saturated regions as highlighted by the red box. In addition to false color generation, Sen12 method exhibits a texture smoothening artifact in the tile region (red dotted zoom box). Similarly, Hu13 method also produces false colors in the saturated regions (blue dotted zoom box). In the second result of Fig. \ref{fig:result2}, as highlighted in blue and red zoom box, Hu13 smoothens details in the cloud region. Similar artifacts by Sen12 and Hu13 are observed in another example from METU dataset \cite{tursun2016objective} in Fig. \ref{fig:result5}. In contrast, the results generated by the proposed method are void of such artifacts. Furthermore, with the help of memory network, it fills in plausible details and generate natural looking results.
    \begin{figure*}[ht]
    \centering
      \includegraphics[width=18cm]{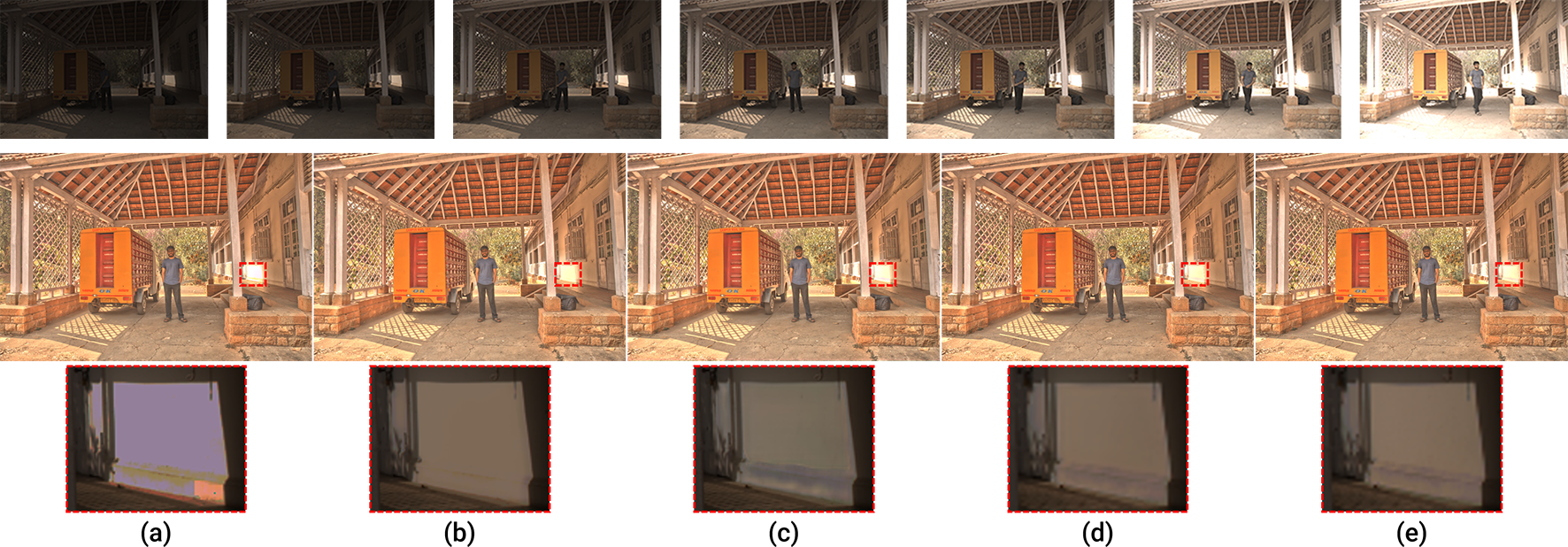}%
      \caption{Our proposed method can easily be extended to fuse arbitrary length sequence by replacing the feature concatenation with Mean+Max fusion strategy followed by Prabhakar \textit{et al.} \cite{prabhakar2019fast}. Top row - the seven varying exposure images captured with Canon EOS-5D Mark III camera in AEB setting with EV-3 to EV+3 bias. Results by (a) Hu13 \cite{hu2013hdr}, (b) Sen12 \cite{sen2012robust}, (c) Prabhakar19 \cite{prabhakar2019fast}, (d) Proposed method, and (e) Ground truth. The zoomed regions shown in the last row are in linear domain to distinguish the artifacts easily.}
      \label{fig:result6}
    \end{figure*}
    
    Kalantari17 method trains a CNN to remove optical flow artifacts in the aligned images. Despite this, some of their results still have flow artifacts. As shown in the black zoomed box of Fig. \ref{fig:result1}, the optical flow artifacts in the region affected by motion and saturation are left uncorrected by their model. Similar artifact is observed in Fig. \ref{fig:result2} and \ref{fig:result4} as well. Prabhakar19 \cite{prabhakar2019fast} method also follows a similar optical flow error correction setup as that of Kalantari17; however, it refines the aligned images using separate networks before fusing them. Despite this, Prabhakar19 method still produces results with visible artifacts in moving saturated regions. This phenomenon is visible in their results in Fig. \ref{fig:result1}, \ref{fig:result2} and \ref{fig:result4}. Both of these methods use optical flow for alignment, followed by fusion with CNN. As evidenced by their results, these two methods find it hard to handle difficult situations because the network treats saturated regions as moving irrespective of motion. 
    By comparison, our proposed method segments images into static and dynamic before fusion, thus avoiding any such confusions.
    
    Unlike Kalantari17 and Prabhakar19, Yan19 and Wu18 train a CNN model to fuse unaligned images. While this approach aids to avoid false color generation or optical flow artifacts, they present challenges of their own. In their method, a single CNN is trained to handle motion, saturation, and fill in saturated details. Hence, in heavily saturated static regions, mistaking them for moving regions, they choose the reference image. It can be observed in the green zoom box of Fig. \ref{fig:result1}. The window is unaffected by motion but saturated in the reference image. Instead of obtaining sharper details from the low exposure image, Yan19 and Wu18 generate blurred details from the reference image. Similar artifacts are observed in the moving saturated regions (blue, magenta and red boxes) of Fig. \ref{fig:result1} and in Fig. \ref{fig:result2}, \ref{fig:result4} and \ref{fig:result5}. Our proposed method is void of such artifacts by employing different fusion sub-networks for static and dynamic parts.
\section{Discussion}
    \label{sec:discussion}
    \subsubsection{Arbitrary size fusion} The individual image features are concatenated for fusion in the proposed approach. Hence, similar to \cite{kalantari2017deep, wu2018deep, yan2019attention}, it requires retraining for a different number of input images. We address this challenge by using the mean + max operation used by Prabhakar19. For $\mathbf{N}$ input images, we obtain $\{\mathbf{E}_N\}$ feature maps using a shared encoder. As the encoder is shared, it can be reused for an arbitrary number of images without retraining. The features are then divided into dynamic and static using a single segmentation network. Instead of concatenating the segmented features, we concatenate the mean and max of all $\{\mathbf{E}_N\}$ features. The fusion networks further process the concatenated dynamic and static features. 

In Table \ref{tab:ablation_expts} A7, we present the quantitative result for this approach on Kalantari17 dataset. Compared with the Prabhakar19 approach, our proposed method offers 1.4 dB improvement in PSNR-T, while offering the same scalability advantage. In Fig. \ref{fig:result6}, we show example results by our approach with seven input images and compare with state-of-the-art scalable methods like Sen12, Hu13, and Prabhakar19. While Hu13 introduces false color in saturated regions, Sen12 over smoothens the necessary details. Whereas, the proposed method has successfully retained the structure with correct color details. 

\subsubsection{Moving region segmentation} The segmentation network can accurately detect moving regions across image pairs. A few qualitative results are shown in Fig. \ref{fig:result3}. Quantitatively, the mean intersection-over-union (IoU) achieved by our segmentation network with respect to annotated masks provided by \cite{prabhakar2020} is 0.686 on 15 validation examples of the \cite{kalantari2017deep} dataset. 

\subsubsection{Running time} Our proposed method can process three input images of 1000$\times$1500 resolution in about 3.31 seconds on an NVIDIA Quadro RTX 6000 GPU. Comparatively, the non-deep learning methods such as Hu13 and Sen12 methods take around 450 and 320 seconds. Among the deep learning methods, Kalantari17 takes up to 57 seconds, and Wu18 takes 6.8 seconds. Unlike Kalantari17 method, our proposed approach avoids the optical flow correction overhead by dealing with dynamic and static regions separately.
\section{Conclusion}
    \label{sec:conc}
    In this paper, we proposed a motion segmentation assisted CNN for HDR deghosting. With the help of predicted motion segmentation maps, the proposed method fuses static and dynamic regions independently. Our method learns different and dedicated fusion rules suited for static and dynamic regions by identifying moving regions. Hence, it is better equipped to handle complex motion and saturation seen in practice. The fused static and dynamic features are combined to generate final high-quality ghost-free HDR results. Furthermore, the proposed memory network accumulates the essential features required to generate plausible textures in the saturated regions. Extensive experimental evaluation exemplifies that the proposed method produces visually appealing results with clear texture and faithful color reconstruction without ghosting artifacts. Additionally, we hope that the presented MEDS dataset will help the research community to advance further in a similar direction.

%





\ifCLASSOPTIONcaptionsoff
  \newpage
\fi



\bibliographystyle{IEEEtran}
\bibliography{main.bib}

%








\end{document}